\pdfoutput=1
\documentclass[conference]{IEEEtran}
\ifCLASSOPTIONcompsoc
  \usepackage[nocompress]{cite}
\else
  \usepackage{cite}
\fi
\ifCLASSINFOpdf
\else
\fi

\usepackage{color}
\usepackage{lipsum}                     
\usepackage{xargs}                      
\usepackage[pdftex,dvipsnames,table]{xcolor}  
\usepackage[colorinlistoftodos,prependcaption, textsize=small]{todonotes}
\newcommandx{\unsure}[2][1=]{\todo[inline,linecolor=red,backgroundcolor=red!25,bordercolor=red,#1]{#2}}
\newcommandx{\change}[2][1=]{\todo[inline,linecolor=blue,backgroundcolor=blue!25,bordercolor=blue,#1]{#2}}
\newcommandx{\info}[2][1=]{\todo[inline,linecolor=OliveGreen,backgroundcolor=OliveGreen!25,bordercolor=OliveGreen,#1]{#2}}
\newcommandx{\improvement}[2][1=]{\todo[inline,linecolor=Plum,backgroundcolor=Plum!25,bordercolor=Plum,#1]{#2}}
\newcommandx{\thiswillnotshow}[2][1=]{\todo[disable,#1]{#2}}

\usepackage{xifthen}
\usepackage[normalem]{ulem}	

\newcounter{cmmnt}
\newcommand{\cmmnt}[4][]{%
	\refstepcounter{cmmnt}%
	{%
		\ifthenelse{\isempty{#1}}{%
			\todo[color=#2!30,linecolor=#2,size=\small]{%
				\sffamily
				\textbf{Comment~\thecmmnt [\uppercase{#3}]:} #4}
		}{%
		\todo[color=#2!30,linecolor=#2,size=\small,#1]{%
			\sffamily
			\textbf{Comment~\thecmmnt [\uppercase{#3}]:} #4}
	}
}}

\newcommand{\instance}[1]{\textsf{#1}}

\usepackage{perpage} 
\MakePerPage{footnote} 

\usepackage[draft]{hyperref}

\usepackage{multirow}

\usepackage{algorithm}
\usepackage{algpseudocode}

\usepackage[nomain,acronym,toc]{glossaries}
\makeglossaries

\usepackage{tikz}
\usepackage{pgfplots}
\usetikzlibrary{shapes,arrows}
\pgfplotsset{grid style={dashed}}

\usepackage{mathtools}
\DeclarePairedDelimiter{\ceil}{\lceil}{\rceil}
\usepackage{fancyhdr}

\newacronym{NHGRI}{\texttt{NHGRI}} {National Human Genome Research Institute} %

\newacronym{CX}{\texttt{CX}}{Cycle Crossover}
\newacronym{OX}{\texttt{OX}}{Order Crossover}
\newacronym{PMX}{\texttt{PMX}}{Partial Mapped Crossover}

\newacronym{GPU}{\texttt{GPU}}{Graphical Processing Unit}
\newacronym{RNA}{\texttt{RNA}}{Ribonucleic Acid}
\newacronym{OP}{\texttt{OP}}{Optimization Problem}
\newacronym{CS}{\texttt{CS}}{Computer Science}
\newacronym{EA}{\texttt{EA}}{Evolutionary Algorithm}
\newacronym{EAs}{\texttt{EAs}}{Evolutionary Algorithms}
\newacronym{GA}{\texttt{GA}}{Genetic Algorithm}
\newacronym{SOP}{\texttt{SOP}}{Single-Objective Problem}
\newacronym{SOPs}{\texttt{SOPs}}{Single-Objective Problems}
\newacronym{MOO}{\texttt{MOO}}{Multi-Objective Optimization}
\newacronym{MOP}{\texttt{MOP}}{Multi-Objective Optimization Problem}
\newacronym{MOPs}{\texttt{MOPs}}{Multi-Objective Optimization Problems}
\newacronym{MA}{\texttt{MA}}{Memetic Algorithm}
\newacronym{MAs}{\texttt{MAs}}{Memetic Algorithms}
\newacronym{QAP}{\texttt{QAP}}{Quadratic Assignment Problem}
\newacronym{bQAP}{\texttt{bQAP}}{bi-objective Quadratic Assignment Problem}
\newacronym{mQAP}{\texttt{mQAP}}{Multi-Objective Quadratic Assignment Problem}
\newacronym{mQAPs}{\texttt{mQAPs}}{Multi-Objective Quadratic Assignment Problems}
\newacronym{II}{\texttt{II}}{Inverted Index}
\newacronym{IIs}{\texttt{IIs}}{Inverted Indexes}
\newacronym{EMOO}{\texttt{EMOO}}{Evolutionary Multi-Objective Optimization}
\newacronym{MOEA}{\texttt{MOEA}}{Multi-Objective Evolutionary Algorithm}
\newacronym{MOEAs}{\texttt{MOEAs}}{Multi-Objective Evolutionary Algorithms}
\newacronym{MOMA}{\texttt{MOMA}}{Multi-Objective Memetic Algorithm}
\newacronym{MOMAs}{\texttt{MOMAs}}{Multi-Objective Memetic Algorithms}

\newacronym{GD}{\texttt{GD}}{Generational Distance}
\newacronym{MGD}{\texttt{GD$^+$}}{Modified Generational Distance}
\newacronym{IGD}{\texttt{IGD}}{Inverted Generational Distance}
\newacronym{MIGD}{\texttt{IGD$^+$}}{Modified Inverted Generational Distance}
\newacronym{DM}{\texttt{DM}}{Delta Measure}
\newacronym{HV}{\texttt{HV}}{Hypervolume}
\newacronym{IHV}{\texttt{IHV}}{Inverted Hypervolume}
\newacronym{SP}{\texttt{SP}}{Spacing}
\newacronym{DV}{\texttt{DV}}{Diversity}
\newacronym{RC}{\texttt{RC}}{Radial Coverage}
\newacronym{MS}{\texttt{MS}}{Maximum Spread}
\newacronym{ER}{\texttt{ER}}{Error Ratio}
\newacronym{SCC}{\texttt{SCC}}{Success Counting}
\newacronym{CM}{\texttt{CM}}{Coverage Metric}

\newacronym{SLS}{\texttt{SLS}}{Stochastic Local Search}
\newacronym{PLS}{\texttt{PLS}}{Pareto Local Search}
\newacronym{TPLS}{\texttt{TPLS}}{Two-Phase Local Search}
\newacronym{RoTS}{\texttt{RoTS}}{Robust Tabu Search}
\newacronym{BLS}{\texttt{BLS}}{Breakout Local Search}

\newacronym{MOACO}{\texttt{MO-ACO}} {Multi-Objective Ant Colony Optimization} %
\newacronym{SPEATWO}{\texttt{SPEA2}} {Strength Pareto Evolutionary Algorithm II} 
\newacronym{MOEAD}{\texttt{MOEA/D}} {Multiobjective Evolutionary Algorithm Based on Decomposition} %
\newacronym{NSGATWO}{\texttt{NSGA-II}} {Nondominated Sorting Genetic Algorithm II} %

\newacronym{ITIM}{\texttt{ITIM}}{Iterative Improvement}
\newacronym{OAV}{\texttt{OAV}}{Organic Air Vehicles}

\newacronym{MOMGAII}{\texttt{MOMGA-II}} {Multi-Objective Fast Messy Genetic Algorithm}%
\newacronym{MOMGAIIa}{\texttt{MOMGA-IIa}} {Multi-Objective Fast Messy Genetic Algorithm a}%

\newacronym{GRASP}{\texttt{GRASP}}{Greedy Randomized Adaptive Search Procedure}
\newacronym{mGRASP}{\texttt{mGRASP/MH}}{Multi-Objective Greedy Randomized Adaptive Search Procedure}
\newacronym{FDC}{\texttt{FDC}}{Fitness Distance Correlation}
\newacronym{MOGWW}{\texttt{MOGWW}}{Multi-Objective Go with Winners}

\newacronym{GWW}{\texttt{GWW}} {Go With Winner} %
\newacronym{mPLS}{\texttt{mPLS}}{Multi-Objective Pareto Local Search}

\newacronym{AIS}{\texttt{AIS}}{Artificial Immune System}

\newacronym{GISMOO}{\texttt{GISMOO}}{Multi-Objective Genetic Immune System} 
\newacronym{PMSMO}{\texttt{PMSMO}}{PMSMO}

\newacronym{TA}{\texttt{TA}}{Transgenetic Algorithm}
\newacronym{TAs}{\texttt{TAs}}{Transgenetic Algorithms}

\newacronym{NSTA}{\texttt{NSTA}}{Non-Dominated Sorting Transgenetic Algorithms}
\newacronym{MOTAD}{\texttt{MOTA/D}}{Multi-Objective Transgenetic Algorithm/Decomposition}  

\newacronym{GNA}{\texttt{GNA}}{Global Network Alignment}
\newacronym{SIMD}{\texttt{SIMD}}{Single Instruction, Multiple Data}

\newacronym{TSP}{\texttt{TSP}}{Traveling Salesman Problem}
\newacronym{ROE}{\texttt{ROE}}{Reduce-Optimize-Expand}

\newacronym{CIBM}{\texttt{CIBM}}{Centre for Bioinformatics, Biomarker Discovery and Information-Based Medicine}

\newacronym{SDP}{\texttt{SDP}}{Semidefinite Programming}

\newacronym{DMLS}{\texttt{DMLS}}{Dominance Based Local Search}

\usepackage{graphicx}

\begin{document}

\newcommand{\citetemp}[1][missing citation]{(\textcolor{red}{#1})} %
\newcommand{\etal}{et al.} 

\newcommand{\approachname}{PasMoQAP}
\newcommand{\myapproach}{\textsc{\approachname}}
\newcommand{\largemyapproach}{\textsc{\large \approachname}}
\newcommand{\hugemyapproach}{\textsc{\huge \approachname}}

\renewcommand{\algorithmicrequire}{\textbf{Input:}}
\renewcommand{\algorithmicensure}{\textbf{Output:}}

\renewcommand\IEEEkeywordsname{Keywords}

%
\title{\approachname: A Parallel Asynchronous Memetic Algorithm for solving the Multi-Objective Quadratic Assignment Problem}

\author{
\IEEEauthorblockN{
	Claudio Sanhueza\IEEEauthorrefmark{1},
	Francia Jim\'enez\IEEEauthorrefmark{1},
	Regina Berretta\IEEEauthorrefmark{1} and 
	Pablo Moscato\IEEEauthorrefmark{1}}

	\IEEEauthorblockA{\IEEEauthorrefmark{1}School of Electrical Engineering  and Computing, The University of Newcastle, Callaghan, NSW, Australia.}
	\IEEEauthorblockA{Emails: \{claudio.sanhuezalobos, francia.jimenezfuentes\}@uon.edu.au, \{regina.berretta, pablo.moscato\}@newcastle.edu.au}
}

\maketitle

\begin{abstract}
Multi-Objective Optimization Problems (\acrshort{MOPs}) have attracted growing attention during the last decades. \acrfull{MOEAs} have been extensively used to address \acrshort{MOPs} because are able to approximate a set of non-dominated high-quality solutions. The \acrfull{mQAP} is a \acrshort{MOP}. The \acrshort{mQAP} is a generalization of the classical \acrshort{QAP} which has been extensively studied, and used in several real-life applications. The \acrshort{mQAP} is defined as having as input several flows between the facilities which generate multiple cost functions that must be optimized simultaneously. In this study, we propose \myapproach, a parallel asynchronous memetic algorithm to solve the \acrlong{mQAP}. \myapproach~is based on an island model that structures the population by creating subpopulations. The memetic algorithm on each island individually evolve a reduced population of solutions, and they asynchronously cooperate by sending selected solutions to the neighboring islands. The experimental results show that our approach significatively outperforms all the island-based variants of the multi-objective evolutionary algorithm \acrshort{NSGATWO}. We show that \myapproach~is a suitable alternative to solve the \acrlong{mQAP}.
\end{abstract}

%
\IEEEpeerreviewmaketitle

\section{Introduction}
\label{sec:introduction}
Mathematical models based on \acrfull{MOPs} have been extensively used to address real-world applications \cite{deb2001multi, marler2004survey}. In \acrshort{MOPs}, the task we face is to simultaneously satisfy multiple and possibly conflicting objectives.
\acrfull{MOEAs} have received especial attention because they are well-suited to tackle a wide variety of \acrshort{MOPs} \cite{zitzler1999multiobjective, zitzler1999evolutionary}.
The \emph{Pareto optimal set}\footnote{The Pareto optimal set corresponds to a set of non-dominated solutions which may be considered as the answer of a Multi-Objective Optimization Problem.} is approximated using \acrshort{MOEAs} by evolving individuals (solutions) typically organized in populations. A population is improved by applying evolutionary operators such as recombination, mutation, and selection. \acrshort{MOEAs} augmented with local search operators have attracted growing attention due to their success when applied for solving complex optimization problems. \acrfull{MAs} are computational methods which combine \acrlong{EAs} and individual local search techniques in order to efficiently address optimization problems  \cite{neri2012handbook}.

Despite their advantages, \acrlong{MAs} may be computationally more expensive. In order to approximate the Pareto optimal set, the algorithms should explore large search spaces. By increasing the number of objectives, the size of the search space grows dramatically and the number of solutions belonging to the Pareto set will increase accordingly. Therefore, longer convergence times are expected and more evaluations must be performed to find the final Pareto set. This is even more critical in real-world optimization problems for which the evaluation of a given solution is computationally expensive. In general, most of the time in \acrshort{MAs} is devoted to the local search procedures which explore user-defined neighborhoods \cite{neri2012handbook}. In the case of combinatorial optimization problems, this neighborhood might be enormous; therefore, an exhaustive exploration is unfeasible.

These drawbacks in \acrshort{MAs} can be addressed in two different ways. First, we can limit the neighborhood exploration by relaxing the acceptance criteria. For example, instead of an exhaustive exploration to find the best solution, we could explore the neighborhood until finding a solution that is better than the current one. Second, an expensive operation when exploring neighborhoods is the computation of the objective functions. In this case, for example, we might define surrogate models of the fitness function or we could compute the cost of a neighbor solution based on the method applied to obtain the neighborhood of the current solution. 

Typically, parallel and distributed schemes are used to speed up the search. Several parallel approaches for multi-objective optimization have been proposed \cite{luna2015parallel}. Including parallelism is not just a technique to accelerate the search process, but also for developing effective search methods. For example, a population can be divided into subpopulations to explore and exploit different regions of the search space, improving the quality of the obtained Pareto set.

In this study, our contributions are: 

\begin{itemize}
\item We propose the \myapproach, an asynchronous parallel multi-objective memetic algorithm based on island models to solve the \acrlong{mQAP}.
\item We evaluate our method using twenty-two benchmark instances. To perform a fair comparison between the techniques, we consider the structure of each instance by varying the correlation between the flows. 
\item Our results show a significant difference between our approach and the state-of-the-art algorithm to solve multi-objective optimization problems. Our method outperforms the parallel variants of the \acrshort{NSGATWO} algorithm at a significance level of 5\% ($\rho < 0.05$).
\end{itemize}

 
The rest of the paper is organized as follows. In Section~\ref{sec:problem_definition}, we describe the \acrlong{mQAP} and we discuss related works. We present \myapproach~in Section~\ref{sec:methodology}. We outline our experiments in Section \ref{sec:experimental_design}, and we present the results in Section~\ref{sec:experiments}. We finish with our conclusions and a discussion about future work in Section \ref{sec:conclusions}.

\section{The Multi-objective Quadratic Assignment Problem}
\label{sec:problem_definition}

The \acrfull{mQAP} is a generalization of the well-known \acrfull{QAP}  \cite{knowles2002towards}. The \acrshort{QAP} was first introduced by Koopmans \etal ~\cite{koopmans1957assignment} and it has been largely studied \cite{loiola2007survey}. The problem belongs to the $\mathcal{NP}$-hard class and it has proven to be difficult even for small instances. It can be presented as the problem of allocating a set of facilities to a set of locations, with the cost being a function of the distance between locations and the flows between facilities. The goal is to assign each facility to a location such that the total cost is minimized. The multi-objective variation considers more than one flow between any pair of facilities. This leads to the joint minimization of several objective functions. Formally, the Multi-Objective Quadratic Assignment Problem can be presented as:

\begin{equation}
\begin{aligned}
	\underset{\pi \in P_n}{\text{minimize}}\;\;
			C(\pi)	 & = \{C^1(\pi), C^2(\pi), ... , C^m(\pi) \} \\
			C^r(\pi)	 & = \sum_{i=1}^{n} \sum_{j=1}^{n} d_{ij}f^r_{\pi(i)\pi(j)}, \: r=1,...,m
\end{aligned}
\end{equation}

\noindent
where $f^r_{\pi(i)\pi(j)}$ represents the flow between the facility $\pi(i)=k$ and $\pi(j)=l$ of the $r$-th flow and $d_{ij}$ is the distance between location $i$ and $j$. $P_n$ represents the set of all permutations $\pi: N \rightarrow N$. The product $d_{ij}\,f^r_{\pi(i)\pi(j)}$ is the $r$-th cost of locating facility $\pi(i)=k$ to the location $i$ and facility $\pi(j)=l$ to the location $j$. The difference between this definition and the original \acrshort{QAP} is that we consider different types of flows ($m$ flows in this case) and we are minimizing each of the cost functions simultaneously.

\section{Related work}

The assumption that the performance of evolutionary algorithms can always be improved in practice by including some kind of local search procedure motivates the study of Garrett and Dasgupta \cite{garrett2006analyzing}. In their work, a memetic algorithm (\acrshort{SPEATWO} + \acrshort{RoTS}) to solve \acrshort{mQAP} is compared against a multi-objective tabu search. The authors found that there is a correlation between the advantage obtained by hybridizing the \acrshort{MOEA} (i.e. \acrshort{SPEATWO} + \acrshort{RoTS}) versus a simple iterated local search algorithm and the distribution of offspring solutions generated via recombination. 


Later, the same authors contributed an empirical comparison of a number of strategies to solve the \acrshort{mQAP} \cite{garrett2009empirical}. Their results show that both the number of objectives and the specific structure of the search space have a strong impact in the performance of different \acrshort{MOMAs}. As the number of objectives is increased, the performance of \acrshort{MOMAs} change, with algorithms that stimulate exploration tending to perform better.


In \cite{gutierrez2011enhanced}, Guti{\'e}rrez et al. presented an enhanced variant of \acrfull{MOGWW} algorithm. The \acrshort{MOGWW} algorithm was hybridized using a modified version of a \acrfull{mPLS} \cite{paquete2004pareto}. The original \acrfull{GWW} \cite{aldous1994go} was improved by including two mechanisms: a proper threshold condition and a non-dominated random walk. With the threshold condition at each stage the solutions are classified in non-dominated fronts. The random walk mechanism allows the approach to maintain a diverse set of solutions, which is a strongly recommended property for an algorithm addressing \acrshort{MOPs}. 


\section{The Parallel Memetic Algorithm to solve the mQAP}
\label{sec:methodology}
In this section, we introduce the proposed \myapproach~to solve the \acrlong{mQAP}. Our method is based on the island model which uses several memetic algorithms to evolve populations of individuals in parallel. The memetic algorithms asynchronously communicate to each other selected solutions as a way to improve the quality of the final Pareto set.

\subsection{Overview}
We design each component of \myapproach~as separate modules which can be easily set by the users. The design allows the creation of flexible island models by just specifying configurations files for each island. Therefore, we can create parallel cooperative strategies, for example, between different algorithms with ease. 
We identify three main components in our approach. First, we define a topology using an island model that organizes how the memetic algorithms communicate to each other. Second, we design our memetic algorithm, using particular recombination and mutation operators. Also, we implemented an archiving strategy to store the non-dominated solutions found during the execution of the algorithm. Finally, we develop a local search operator which enhances the evolution process of our algorithm. In the following sections, we discuss the details of each component involved in our algorithm.

\subsection{The Parallel Island Model}
\begin{figure}[t]
	\centering
	\includegraphics[width=0.4\textwidth]{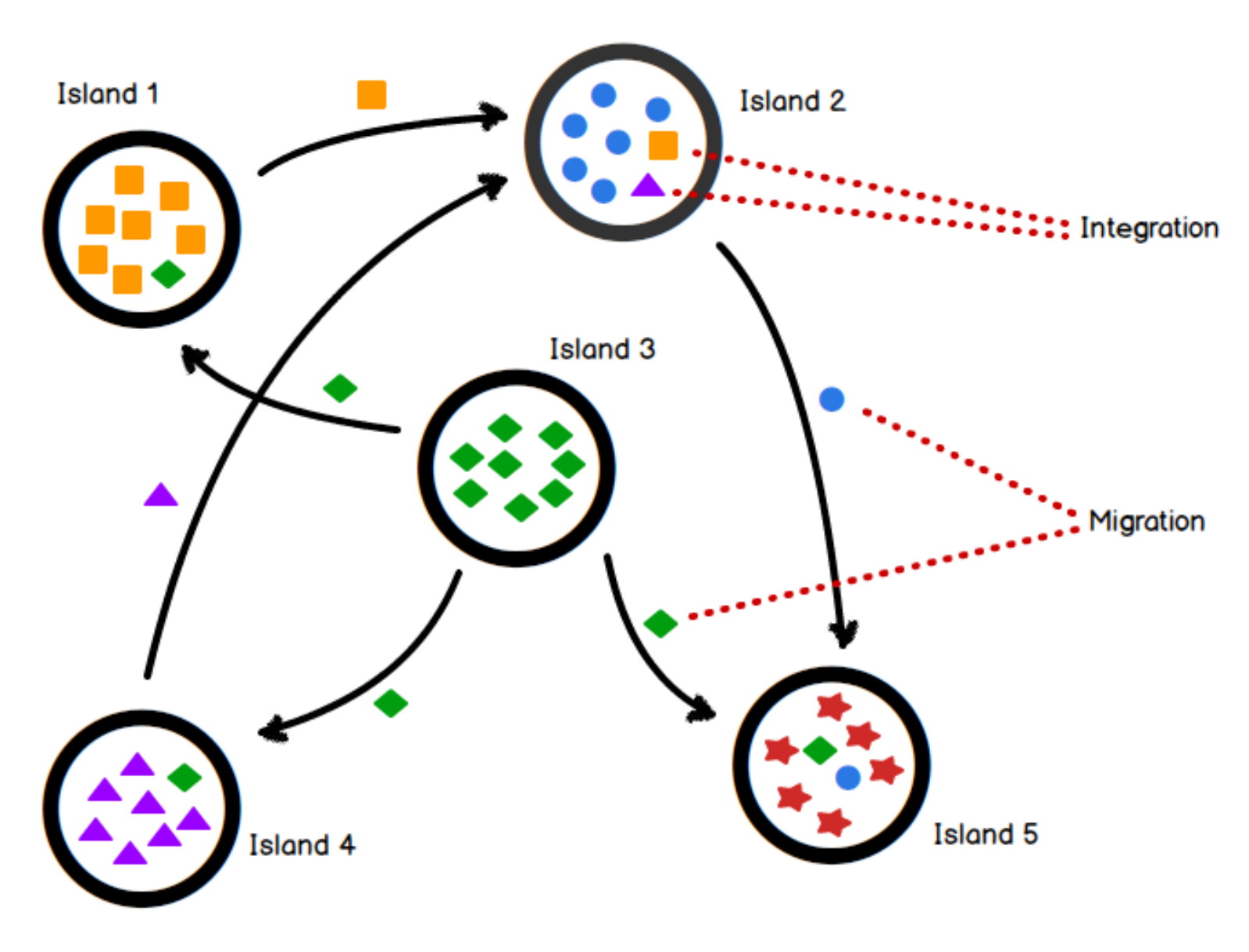}
	\caption{A representation of an island model. Five islands represented by the circumferences evolves solutions (shapes) and periodically and asynchronously they communicate selected solutions to the neighboring islands.}
	\label{background:fig:island_model}
\end{figure}

The parallel island model belongs to the \emph{self-contained parallel cooperation} approach described in \cite{talbi2008parallel}. In the self-contained parallel cooperation each processor executes an independent algorithm using a sub-population. Typically, the cooperation between the algorithms is performed by selecting and sending migrants to the other processors. These algorithms are suitable to address problems with large search space. The population is divided into several subpopulations, called \emph{islands} or \emph{demes}. Each island computes an algorithm during a given period called \emph{epoch}. After the epoch is reached, a \emph{selection criteria} is applied to choose the individuals that will be migrated between islands. A \emph{migration path} is defined by using a \emph{island-network topology} which determines how the individual can be moved to other islands. Moreover, each island defines its \emph{integration strategy}. The integration strategy is applied every time new individuals arrive from a migration procedure, and it determines if the immigrants should be included in the current island. An example of the island model is depicted in \autoref{background:fig:island_model}.

\subsection{Our approach: \largemyapproach}

\myapproach~is based on ideas from the island model, where the overall population is split into subpopulations, called \emph{islands}. Each island evolves independently using a \acrlong{MA}. In our method, the islands are logically connected to each other, asynchronously migrating solutions at a fixed rate. The immigrant solutions received by the islands are evaluated against the members of the current population. The final goal of our method is to solve the \acrlong{mQAP} by creating a cooperation model to improve the quality of the solutions.

\begin{algorithm}[t]
    \caption{Pseudocode of \myapproach~in a specific island}
    \label{methodology:algo:island}
    
    \begin{algorithmic}[1] 
	    \Require population size $n_p$, set of neighboring islands $I$, island identification $i$, number of migrants $m$, epoch $e$, number of generations $g_{max}$, probability of crossover $pb_c$, probability of mutation $pb_m$
	    \Ensure Final subset of non-dominated solutions $\mathcal{P}^{*}_i$
	        \State $\mathcal{P}^{*}_i \gets \emptyset$
	        \State $P \gets \Call{RandomInit}{n_p}$
	        \Comment{Initialize subpopulation $P$ at random}
            \State $\mathcal{P}^{*}_i \gets \mathcal{P}^{*}_i \cup P $ \Comment{Initial set of non-dominated solutions}
            \State $g_c \gets 1$
            \While{ $g_c \leq g_{max}$ }
	            \State $parents \gets \Call{ParentsTourSelection}{P}$
	            \State $ O \gets \Call{CycleCrossover}{parents, pb_c}$
           		\State $O \gets \Call{SwapMutation}{O, pb_m}$
           		\State $\mathcal{P}^{*}_i \gets \mathcal{P}^{*}_i \cup O $
           		\State $P \gets \Call{DomBasedLocalSearch}{\mathcal{P}^{*}_i}$
	            \State $ P \gets \Call{DomDepthFitAssignment}{P}$
	            \State $ P \gets \Call{FxFCrowdingDivAssignment}{P}$
	            \State $M \gets \Call{CheckMigrants}{I}$ 
	            \Comment{Checking the arrival of migrants from neighboring islands}
	            \State $\mathcal{P}^{*}_i \gets \mathcal{P}^{*}_i \cup P \cup M$
	            \If{$g_c~\%~e == 0$}
		            \State $S \gets \Call{TournamentSelection}{\mathcal{P}^{*}_i, m}$
		            \State $\Call{SendSolutions}{S, I}$
		            \Comment{Sending solutions to the neighboring islands}
		        \EndIf
	            \State $P \gets \Call{ElitistIntegration}{P \cup M}$
	            \State $g_c++$              
            \EndWhile\label{moma_while}
            \State \textbf{return} $\mathcal{P}^{*}_i$ \Comment{Return the subset of non-dominated solutions for the island $i$}
    \end{algorithmic}
\end{algorithm}

In Algorithm \ref{methodology:algo:island}, we present the pseudocode of \myapproach~in a specific island. First, the external archive of the $i$-th island $\mathcal{P}^{*}_i$ is initialized and the subpopulation $P$ is initiated at random. The main components of our memetic algorithm are executed during a fixed number of generations $g_{max}$. The \textsc{CycleCrossover} is executed with probability $pb_c$ and the \textsc{SwapMutation} with probability $pb_m$. In line 9, the archive is updated with the offspring $O$, and later the archive is used as input in our local search procedure called \textsc{DomBasedLocalSearch} (line 10). Once the local search procedure is executed, the algorithm computes the fitness (\textsc{DomDepthFitAssignment}) and diversity (\textsc{FxFCrowdingDivAssignment}) measures for each individual. An island receives, without blocking, the migrants $M$ from the neighboring islands $I$, using the function \textsc{CheckMigrants}. In line 14, the external archive is updated, including the current subpopulation as well as the immigrants. Everytime the epoch ($e$) is reached, the candidates $S$ to be migrated are selected with the \textsc{TournamentSelection}, and a copy of them is sent to the destination islands using \textsc{SendSolutions}. In line 19, the subpopulation is updated, replacing some individuals with the immigrants. We employ \textsc{ElitistIntegration}, in which the current population and the immigrants are combined, and then they are sorted using the Pareto dominance relation. This elitist method keeps the currently best solutions for the next generation, assuring proximity to the optimal Pareto front. Finally, the algorithm collects and mixes the external archives of each island to obtain the global Pareto set.

In the following sections, we present the details of each component of \myapproach.

\subsubsection{Island model}
We design \myapproach~with independent modules which communicate to each other using ad-hoc interfaces to create a topology of islands. We develop in this way to facilitate re-usability of our implementations. Therefore, we can easily perform our experiments, and we can extend our approach for further studies.

\paragraph{Topology}
We analyze \myapproach's performance with a range of islands. In our experiments, we test the performance of 5, 8, 11, 16, and 21 memetic algorithms running in parallel. To setup a comparable scenario, we implement a simple communication topology called \emph{complete} where each island is connected to each other. The algorithms asynchronously communicate, sharing promising individuals at a fixed rate.

\paragraph{Migration}
An island in \myapproach~sends two promising solutions ($m=2$) every five generations ($e=5$). The migration process is performed using two steps. First, the migrants are selected using a selection operator. In particular, we use deterministic tournament \cite{miller1995genetic} between the solutions (\textsc{TournamentSelection}). Second, the algorithm compares the solutions to apply the selection criteria. A \emph{comparator} is an operator that is used to compare the quality between two solutions (i.e. to determine which one is better than the other). Our approach uses a comparator that first compares the fitness values of two solutions (i.e. convergence quality), and in the case of a tie, it compares the diversity values. This comparator is called \emph{FitnessThenDiversity}. We present the details of the computation of the fitness and diversity values in the following sections. 

\paragraph{Integration}
Each time an island receives immigrant solutions, the algorithm must decide whether to integrate them into the current population. We choose to use an \textsc{ElitistIntegration} strategy that consists of keeping the best individuals in the population. The procedure updates the population by adding the migrant solutions, sorting them with the \emph{FitnessThenDiversity} comparator and resizing the whole population obtained.

\paragraph{Fitness assignment}
Different schemes exist to assign fitness values to the individuals contained in the population at each generation. We must note that this measure aims to represent the quality of an individual regarding convergence (according to the other solutions in the population) and it should not be confused with the objective values of a particular solution. \myapproach~uses the dominance based fitness assignment strategy called \emph{Dominance Depth fitness assignment} \cite{srinivas1994muiltiobjective, deb2002fast}. In \textsc{DomDepthFitAssignment}, given a population of solutions, the procedure computes the ranking of a solution $p$ by counting the number of solutions $C_p$ (\emph{domination count}) that dominate $p$, and $S_p$ a set of solutions that the solution $p$ dominates. Thus, the procedure creates groups of non-dominated solution where the first group will have their domination values as zero. Next, for each solution $p$ with $C_p = 0$, the method visits each member $q \in S_p$ and reduces its domination count by one. In doing so, if for any member $q$ the domination count becomes zero the member is saved in a new list $Q$. These members belong to the second non-dominated group. The procedure is repeated with the members stored in $Q$ for identifying the third group. The process continues until all the groups are found. At the end, each solution will have a ranking value that indicates the position in the ranking of a solution.

\paragraph{Diversity assignment}
Likewise, different schemes exist to assign diversity measures to the individuals contained in the population at each generation. The measure aims to represent the quality of an individual in terms of diversity. \myapproach~uses a \emph{front-by-front crowding diversity assignment} strategy which in Algorithm \ref{methodology:algo:island} we called \textsc{FxFCrowdingDivAssignment}. The computation of this measure requires sorting the population according to each objective function value in ascending order of magnitude. After that, for each objective function, the boundary solutions (i.e. solutions with smallest and largest objective values) are assigned with an infinite distance value. All other intermediate solutions are assigned a distance value equal to the normalized absolute difference in the function values of two adjacent solutions. The computation is repeated in each objective. The overall crowding diversity measure is calculated as the sum of individual distance values corresponding to each objective. The method normalizes each objective value between 0 and 1 before calculating the diversity measure.

\subsubsection{Memetic Algorithm}
In the following sections, we present the details of each component in the memetic algorithm. First, we discuss the details of the evolutionary aspect of our approach. Later, we explain the details of our local search procedure. In particular, we describe the current set selection strategy, the neighborhood exploration strategy, and the stopping criteria.

\paragraph{Population} The population on each experiment with 5, 8, 11 islands was defined as $n_p = \ceil{100/N_i}$ where $N_i$ represents the number of islands. We also perform experiment on large island models each one with 16 and 21 islands. To avoid small populations in this case, we define the population size as the mean of our previous experiment i.e. $n_p=13$.

\paragraph{Recombination and mutation}
In this study, we test several recombination operators such as \emph{order crossover} (\acrshort{OX}) \cite{golberg1989genetic}, \emph{cycle crossover} \cite{oliver1987study} (\acrshort{CX}), and \emph{partial mapped crossover} (\acrshort{PMX}) \cite{sivanandam2008introduction}. The best results were obtained with the \emph{cycle crossover}. The \textsc{CycleCrossover} preserves the position of the elements in the parent by identifying \emph{cycles} between the two parents. A cycle corresponds to a sequence of elements which is obtained by alternately visiting elements of each parent. The method works as follows. First, randomly select a parent and a cycle starting point. Next, the element at the cycle starting point of the selected parent is inherited by the child. Later, the element in the same position in the other parent cannot then be placed in this position. Then, its position is found in the selected parent and is inherited from that position by the child. Repeat the process until the cycle is completed by finding the first item in the non-selected parent. To select the two parents for applying the recombination operator we use the tournament selection procedure.

To form the children, the cycles are copied from the respective parents in alternating order, i.e. in cycle 1, the elements of parent 1 are copied to child 1, while in cycle 2 the elements of parent 1 are copied to child 2, and so on. It is important to note that cycle crossover always keeps the position of items from one parent or the other without any alteration. We applied the crossover operator with a probability of $pb_c=0.9$. An example of the operator is depicted in \autoref{methodology:fig:cycle_crossover}.
  
To explore the search space of solutions, we use the \textsc{SwapMutation} operator with probability $pb_m=0.01$. The operator consists of exchanging two randomly selected elements so that each occupies the location formerly occupied by the other.
  
\begin{figure}[t]
		\begin{center}
		\resizebox{\textwidth/2}{!}{%
			\begin{tabular}{ |c|l|c|l| }
				\hline
				\textbf{Cycle} & \textbf{Parents} & \textbf{Elements} & \textbf{Children} \\ 
				\hline
				1 & P1: \colorbox{BurntOrange}{8}  4  7  3  6  2  5  1  \colorbox{BurntOrange}{9  0} 	& 8,  0,  9 		& C1: \colorbox{BurntOrange}{8}  -  -  -  -  -  -  -  \colorbox{BurntOrange}{9 0} \\ 
				  & P2: \colorbox{Apricot}{0}  1  2  3  4  5  6  7  \colorbox{Apricot}{8  9} 	&   			& C2: \colorbox{Apricot}{0}  -  -  -  -  -  -	-  \colorbox{Apricot}{8 9} \\ 
				\hline
				2 & P1: 8  \colorbox{Green}{4  7}  3  \colorbox{Green}{6  2  5  1}  9  0 		& 4, 1, 7, 2, 5, 6 			& C1: \colorbox{BurntOrange}{8} \colorbox{LimeGreen}{1 2} - \colorbox{LimeGreen}{4 5 6 7} \colorbox{BurntOrange}{9 0} \\ 
				  & P2: 0  \colorbox{LimeGreen}{1  2}  3  \colorbox{LimeGreen}{4  5  6  7}  8  9 		&   			& C2: \colorbox{Apricot}{0} \colorbox{Green}{4 7} - \colorbox{Green}{6 2 5 1} \colorbox{Apricot}{8 9} \\ 
				\hline
				3 & P1: 8  4  7  \colorbox{NavyBlue}{3}  6  2  5  1  9  0 						& 3 			& C1: \colorbox{BurntOrange}{8} \colorbox{LimeGreen}{1 2} \colorbox{NavyBlue}{3} \colorbox{LimeGreen}{4 5 6 7} \colorbox{BurntOrange}{9 0} \\  
				  & P2: 0  1  2  \colorbox{Cyan}{3}  4  5  6  7  8  9 						&  				& C2: \colorbox{Apricot}{0} \colorbox{Green}{4 7} \colorbox{Cyan}{3} \colorbox{Green}{6 2 5 1} \colorbox{Apricot}{8 9} \\
				
				\hline
			\end{tabular}}
		\end{center}
	\caption{A \acrfull{CX} operator example. We highlight the elements to indicate how the values from the parents are copied to the children in every cycle.}
	\label{methodology:fig:cycle_crossover}
\end{figure}



\paragraph{Archive}
\myapproach~uses an external archive during the optimization process. An archive is a secondary population that stores non-dominated solutions found since the beginning of the optimization process. It aims first at preserving these solutions by updating the archive at each generation. But, it is also possible to include archive's members during the selection phase of the evolutionary algorithm, to save the archive's objective vectors into a file every generation, and/or to compute performance metrics on this archive. We must keep in mind that the default dominance relation used to update the archive is the Pareto dominance relation, but other relations (typically relaxations of the Pareto dominance) can be specified. In particular, our approach employs an archive with a limited number of solutions to avoid an exponential grow of the archive's size. The new non-dominated solutions are included in the archive according to the Pareto dominance relation.

\subsubsection{Local Search}
After applying the evolutionary operators, our approach runs a \emph{Dominance Based Local Search} (\acrshort{DMLS}) algorithm \cite{liefooghe2009study}. A \acrshort{DMLS} algorithm defines both problem-related and problem-independent modules. Next, we discuss the current set selection strategy, the neighborhood exploration strategy, and the stopping criteria. These strategies belong to the problem-independent modules. We present the details of the local search procedure used in \myapproach~in Algorithm \ref{methodology:algo:ls}.


\begin{algorithm}[t]
	\caption{Pseudocode of the \textsc{DomBasedLocalSearch} procedure}
	\label{methodology:algo:ls}
	
	\begin{algorithmic}[1] 
		\Require archive of the island $\mathcal{P}^{*}_i$
		\Ensure A subpopulation $P$
		\State Mark $\pi$ unvisited $\forall~\pi \in \mathcal{P}^{*}_i$
		\State $P \gets \mathcal{P}^{*}_i$
		\State $t_0 \gets \Call{WallTime}{clock}$
		\State $t_c \gets 0$
		\While{($t_c < t_{max}$) or ($\mathcal{P}^{*}_i = \emptyset$)  }
			\State $\pi \gets \Call{SelectSolution}{\mathcal{P}^{*}_i}$
			\For{\textbf{each} $\pi' \in \Call{OrderSwapExploration}{\pi}$ }
			\If {$\pi'$ \emph{dominates} $\pi$}
				\State $\pi'$.visited $\gets$ false
				\State $P \gets P \cup \pi'$
				\State $\pi$.visited $\gets$ true
				\State \textbf{break}
			\EndIf
			\EndFor
			
			\State $\mathcal{P}^{*}_i \gets \Call{GetUnvisited}{P \cup \mathcal{P}^{*}_i}$
			\State $t_c \gets \Call{WallTime}{clock} - t_0$
		\EndWhile
		\State \textbf{return} $P$
	\end{algorithmic}
\end{algorithm}

\paragraph{Current set selection}
The first step of a local search procedure deals with the selection of a set of solutions from which the neighborhood will be explored. In general, a DMLS model could apply two strategies: \emph{exhaustive selection} and \emph{partial selection}. In the former, the whole set of solutions stored in the archive is selected. On the other hand, only a subset of solutions is chosen in a partial selection. Such a set may be selected at random, or also with respect to a diversity measure. 
The local search procedure used in \myapproach~applies an exhaustive selection of individuals. During each iteration of the local search procedure, an unvisited solution is selected at random from the archive using the \textsc{SelectSolution} procedure. This \emph{current set} is the starting point for the next step in our local search procedure, the neighborhood exploration strategy.

\paragraph{Neighborhood exploration}
From the current set, candidate solutions must be generated using a neighborhood structure. The neighborhood is obtained by applying a local transformation to every solution belonging to the current set. 
We use the swap operator to define our neighborhood. This operator interchanges two selected facilities from their original positions. We use this operator because we can compute in $\mathcal{O}(m \times n)$ the objectives functions of a new solution by just computing the difference between the original and a new solution.

Given the solution $\pi$ and the locations $i$ and $j$ of the elements to be exchanged, the difference $\delta$ in the $r$-th objective function value when exchanging the elements $\pi(i)$ and $\pi(j)$ is computed as follows \cite{taillard1995comparison}.

\begin{equation}
\begin{aligned}
	\delta^{r}(\pi,i,j) = 	(d_{ii} - d_{jj}) ( f_{\pi(j) \pi(j)}^{r} - f_{\pi(i) \pi(i)}^{r} ) + \\ 
					(d_{ij} - d_{ji}) ( f_{\pi(j) \pi(i)}^{r} - f_{\pi(i) \pi(j)}^{r} ) + \\
					\sum_{ \substack{k=1 \\ k \neq i,j} }^{n} 
					\Big((d_{ki} - d_{kj}) (f_{\pi(k) \pi(j)}^{r} - f_{\pi(k) \pi(i)}^{r}) + \\ 
					(d_{ik} - d_{jk}) (f_{\pi(j) \pi(k)}^{r} - f_{\pi(i) \pi(k)}^{r}) \Big) \\
\end{aligned}
 \end{equation}

%

Our local search procedure uses an \textsc{OrderSwapExploration} strategy. In this case, the first selected pair of facilities are located in the positions 0 and 1, the second pair of facilities will be located in positions 0 and 2, and so on.

In general, we can identify two classes of neighborhood exploration. An \emph{exhaustive neighborhood exploration} generates every possible move and the neighboring solutions are all added to the candidate set. A \emph{partial neighborhood exploration} generates a subset of moves. The \myapproach~local search procedure uses the \emph{FirstImproving} strategy which belongs to the partial exploration class. The method explores the neighborhood by transforming the current solution $\pi$, and incrementally computes the objective functions of the transformed solution $\pi'$. If the transformation generates a better solution ($\pi'$ \emph{dominates} $\pi$, line 8 in Algorithm \ref{methodology:algo:ls}), the method accepts the new solutions to be included in the subpopulation, and finishes the exploration for the solution $\pi$. Finally, the subpopulation is updated with the solutions of the archive, and the procedure keeps only the unvisited solutions for the next iteration  (\textsc{GetUnvisited}).

\paragraph{Stopping criteria}
In \myapproach, the local search procedure is allowed to run during 5 seconds ($t_{max}$) or until all the solutions in the archive are visited ($\mathcal{P}^{*}_i = \emptyset$). In this way, we avoid the local search monopolize the execution of the algorithm during long periods.

\section{Experimental design}
\label{sec:experimental_design}
In this section, we provide an overview of the experimental design used in our study to evaluate \myapproach. We present the set of instances which we use to evaluate our method, the baseline algorithm that we use to perform the comparisons, and the metric that we compute to measure convergence of the different approaches.

\subsection{Instances} 

We evaluate \myapproach~with a set of twenty-two 60 facility benchmark instances defined in \cite{garrett2009empirical}\footnote{\url{https://github.com/csanhuezalobos/gar60}}. We re-create the instances using the instance generator presented in \cite{knowles2003instance} with the parameters that are shown in \autoref{experiment:table:instances}. The set of instances includes \acrlong{mQAPs} instances with two, three and four objectives. We consider the problems' structure by varying the \emph{correlation} between flow matrices. In \emph{uniform} instances, the flow values are uniformly distributed across the matrices; hence, the instances represent less realistic \acrshort{mQAPs}. On the other hand, the \emph{real-like} instances include sparsity during the generation, representing more realistic instances of the \acrshort{mQAP} \cite{knowles2003instance}. We use this variety of instances to ensure that the algorithms are not being selected because of good performances on a small set of very similar benchmark instances.

\begin{table}
\tiny
\caption{The multi-objective \acrshort{QAP} benchmark instances.} \label{experiment:table:instances}
\begin{center}
\resizebox{\textwidth/2}{!}{%
\begin{tabular}{ clccc }
\hline
\textbf{Objectives} & \textbf{Instance} & \textbf{Size} & \textbf{Type} & \textbf{Correlation} \\ \hline
\multirow{10}{*}{2} & \instance{Gar60-2fl-1uni} & 60 & uniform & -0.3 \\
 					& \instance{Gar60-2fl-2uni} & 60 & uniform & 0 \\
 					& \instance{Gar60-2fl-3uni} & 60 & uniform & 0.3 \\
 					& \instance{Gar60-2fl-4uni} & 60 & uniform & -0.8 \\
 					& \instance{Gar60-2fl-5uni} & 60 & uniform & 0.8 \\
 					& \instance{Gar60-2fl-1rl} & 60 & real-like & -0.3 \\
 					& \instance{Gar60-2fl-2rl} & 60 & real-like & 0 \\
 					& \instance{Gar60-2fl-3rl} & 60 & real-like & 0.3 \\
 					& \instance{Gar60-2fl-4rl} & 60 & real-like & -0.8 \\
 					& \instance{Gar60-2fl-5rl} & 60 & real-like & 0.8 \\
 					\hline
\multirow{6}{*}{3} 	& \instance{Gar60-3fl-1uni} & 60 & uniform & 0 \\
 					& \instance{Gar60-3fl-2uni} & 60 & uniform & -0.5 \\
 					& \instance{Gar60-3fl-3uni} & 60 & uniform & 0.5 \\
 					& \instance{Gar60-3fl-1rl} & 60 & real-like & 0 \\
 					& \instance{Gar60-3fl-2rl} & 60 & real-like & -0.5 \\
 					& \instance{Gar60-3fl-3rl} & 60 & real-like & 0.5 \\
					\hline 
\multirow{6}{*}{4} 	& \instance{Gar60-4fl-1uni} & 60 & uniform & 0 \\
 					& \instance{Gar60-4fl-2uni} & 60 & uniform & -0.5 \\
 					& \instance{Gar60-4fl-3uni} & 60 & uniform & 0.5 \\
 					& \instance{Gar60-4fl-1rl} & 60 & real-like & 0 \\
 					& \instance{Gar60-4fl-2rl} & 60 & real-like & -0.5 \\
 					& \instance{Gar60-4fl-3rl} & 60 & real-like & 0.5 \\
					\hline
\end{tabular}} 
\end{center}
\end{table}

\subsection{Baseline}

To evaluate \myapproach, we compare it against a well-known multi-objective evolutionary algorithm called \acrshort{NSGATWO} \cite{deb2002fast}. This algorithm has been extensively used for tackling multi-objective optimization problems; therefore, it is considered the state-of-the-art algorithm \cite{pilat2010evolutionary,zhou2011multiobjective}. To be fair in our comparisons, we implemented a baseline island models using the \acrshort{NSGATWO} algorithm on each island.

\subsection{Performance assesment}

To quantify the convergence of our method we used the \emph{hypervolume} indicator proposed by Zitzler \etal \cite{zitzler1999multiobjective, zitzler1999evolutionary}. The hypervolume quantifies the volume, in the objective space, of the space dominated by the obtained non-dominated Pareto optimal solutions bounded by a \emph{reference point}. The hypervolume has been proven to be Pareto compliant \cite{fonseca2005tutorial, zitzler2007hypervolume}, i.e., it does not contradict the order induced by the Pareto dominance relation. Therefore, higher values of the hypervolume indicator mean better performances.


\section{Computational results}
\label{sec:experiments}

\begin{table*}[t]
	\centering
	\caption{The mean of the normalized hypervolume indicator. We evaluate five \acrshort{NSGATWO} island models using 5, 8, 11, 16, and 21 islands. We also evaluate five \myapproach~island models using the same number of islands. We highlight the best results per instance.}
	\label{experiments:fig:hypervolume}
	\begin{tabular}{|l|ccccc|ccccc|}
		\hline
		& \multicolumn{5}{c|}{NSGA-II} & \multicolumn{5}{c|}{\myapproach} \\ \hline
		Instances & \multicolumn{1}{c}{5} & \multicolumn{1}{c}{8} & \multicolumn{1}{c}{11} & \multicolumn{1}{c}{16} & \multicolumn{1}{c|}{21} & \multicolumn{1}{c}{5} & \multicolumn{1}{c}{8} & \multicolumn{1}{c}{11} & \multicolumn{1}{c}{16} & \multicolumn{1}{c|}{21} \\ \hline
		\instance{Gar60-2fl-1uni} & 0.8659 & 0.8745 & 0.8399 & 0.8570 & 0.8838 & 0.8790 & 0.8913 & \cellcolor{blue!25}0.9128 & 0.8714 & 0.9080 \\
		\instance{Gar60-2fl-2uni} & 0.8529 & 0.8388 & 0.8421 & 0.8407 & 0.8551 & 0.8519 & 0.8749 & 0.8961 & 0.8587 & \cellcolor{blue!25}0.9127 \\
		\instance{Gar60-2fl-3uni} & 0.8130 & 0.7893 & 0.8094 & 0.8136 & 0.8025 & 0.8415 & 0.8107 & 0.8673 & 0.8155 & \cellcolor{blue!25}0.8868 \\
		\instance{Gar60-2fl-4uni} & 0.7846 & 0.7788 & 0.7656 & 0.7543 & 0.7613 & 0.8043 & 0.7944 & 0.8001 & 0.7677 & \cellcolor{blue!25}0.8040 \\
		\instance{Gar60-2fl-5uni} & 0.3803 & 0.4031 & 0.3988 & 0.4472 & 0.4993 & 0.5037 & 0.3529 & 0.5188 & 0.3235 & \cellcolor{blue!25}0.5255 \\
		\instance{Gar60-2fl-1rl} & 0.8952 & 0.8294 & 0.8598 & 0.9022 & 0.8757 & 0.9180 & 0.9169 & \cellcolor{blue!25}0.9524 & 0.8697 & 0.9393 \\
		\instance{Gar60-2fl-2rl} & 0.8860 & 0.8657 & 0.8595 & 0.8307 & 0.8449 & 0.9259 & 0.9144 & \cellcolor{blue!25}0.9473 & 0.8685 & 0.9419 \\
		\instance{Gar60-2fl-3rl} & 0.9010 & 0.8705 & 0.8852 & 0.8607 & 0.8852 & 0.8945 & 0.8812 & \cellcolor{blue!25}0.9209 & 0.8797 & 0.8996 \\
		\instance{Gar60-2fl-4rl} & 0.8908 & 0.8700 & 0.8681 & 0.8840 & 0.9035 & 0.9307 & 0.9202 & \cellcolor{blue!25}0.9428 & 0.9090 & 0.9325 \\
		\instance{Gar60-2fl-5rl} & 0.7734 & 0.8264 & 0.7456 & 0.7725 & 0.7241 & \cellcolor{blue!25}0.8740 & 0.8376 & 0.8704 & 0.8032 & 0.8253 \\ \hline
		\instance{Gar60-3fl-1uni} & 0.8493 & 0.8332 & 0.8694 & \cellcolor{blue!25}0.9090 & 0.9039 & 0.8550 & 0.8324 & 0.8910 & 0.8963 & 0.8843 \\
		\instance{Gar60-3fl-2uni} & 0.7633 & 0.7539 & 0.7703 & 0.7190 & 0.7308 & \cellcolor{blue!25}0.8010 & 0.7838 & 0.7963 & 0.7398 & 0.7737 \\
		\instance{Gar60-3fl-3uni} & 0.8004 & 0.7531 & 0.7662 & 0.7789 & 0.7532 & 0.8494 & 0.8022 & \cellcolor{blue!25}0.8730 & 0.7961 & 0.7268 \\
		\instance{Gar60-3fl-1rl} & 0.8802 & 0.8565 & 0.9272 & \cellcolor{blue!25}0.9513 & 0.9194 & 0.8855 & 0.9009 & 0.9157 & 0.9460 & 0.9476 \\
		\instance{Gar60-3fl-2rl} & 0.8537 & 0.8506 & 0.8872 & \cellcolor{blue!25}0.9324 & 0.8648 & 0.8703 & 0.8991 & 0.9006 & 0.9275 & 0.8744 \\
		\instance{Gar60-3fl-3rl} & 0.8579 & 0.8687 & 0.8719 & 0.8858 & \cellcolor{blue!25}0.9107 & 0.8962 & 0.8876 & 0.9048 & \cellcolor{blue!25}0.9107 & 0.8932 \\ \hline
		\instance{Gar60-4fl-1uni} & 0.7817 & 0.8123 & 0.8726 & 0.8252 & 0.8000 & 0.8134 & 0.8111 & \cellcolor{blue!25}0.9083 & 0.8556 & 0.8251 \\
		\instance{Gar60-4fl-2uni} & 0.6892 & \cellcolor{blue!25}0.7237 & 0.6920 & 0.6170 & 0.6133 & 0.7019 & 0.7184 & 0.6679 & 0.6480 & 0.6465 \\
		\instance{Gar60-4fl-3uni} & 0.7920 & 0.7764 & 0.7916 & 0.7893 & 0.7993 & 0.8017 & 0.7642 & 0.8729 & 0.7938 & \cellcolor{blue!25}0.8995 \\
		\instance{Gar60-4fl-1rl} & 0.7264 & 0.8759 & 0.9181 & 0.8466 & 0.8488 & 0.8221 & 0.8684 & \cellcolor{blue!25}0.9200 & 0.9018 & 0.8692 \\
		\instance{Gar60-4fl-2rl} & 0.8083 & 0.8599 & \cellcolor{blue!25}0.9145 & 0.8354 & 0.8091 & 0.8338 & 0.8553 & 0.9131 & 0.8728 & 0.8640 \\
		\instance{Gar60-4fl-3rl} & 0.8015 & 0.8569 & 0.8914 & 0.8842 & 0.8557 & 0.8248 & 0.8275 & \cellcolor{blue!25}0.9201 & 0.8783 & 0.8823 \\
		\hline
	\end{tabular}
\end{table*}

 We implement \myapproach~in C++ using the framework ParadisEO \cite{cahon2004paradiseo,liefooghe2010paradiseo}.
 The experiments were performed in the University of Newcastle's Research Compute Grid that contains a Cluster of 32 nodes Intel\textregistered~Xeon\textregistered~CPU E5-2698 v3 @ 2.30 GHz x 32 with 128 GB of RAM.
 
 We execute 30 independent trials for each benchmark instance. The algorithms were allowed to run for 100 generations or during 300 seconds (5 min.). Then, we get a reference point per instance from a global Pareto set that we compute by combining the results of each trial. We use each reference point to measure the hypervolume in each computed Pareto set. Finally, we report the average normalized hypervolume metric. All the objectives values were normalized between 0 and 1 before performing the hypervolume computation.
  
 In our experiments, we evaluate ten different island-based algorithms. Five correspond to the baseline \acrshort{NSGATWO} island model using 5, 8, 11, 16, and 21 islands and five correspond to \myapproach~with the same number of islands. We highlight the best hypervolume indicators for each instance in \autoref{experiments:fig:hypervolume}. Comparing the results between island models with the same number of islands, we observe that \myapproach~consistently outperforms the  baseline \acrshort{NSGATWO} method in terms of the hypervolume with a few exceptions where \acrshort{NSGATWO} obtain better results. 
 
 We apply the Wilcoxon rank sum test to the mean hypervolume indicator values to determine if there are significant differences between the algorithms at a significance level of 5\%. First, we perform a pairwise comparison between \acrshort{NSGATWO} and \myapproach~when they use the same number of islands. Our results suggest that there are no significant differences between the algorithms with 5, 8, 16, and 21 islands. We found significant differences ($\rho=0.0074$) between the approaches with 11 islands. Second, we use \myapproach~with 11 islands to compare against the rest algorithms which use a different number of islands. Our statistical tests show that \myapproach~with 11 islands significantly outperforms the whole set of \acrshort{NSGATWO} island models. Moreover, \myapproach~with 11 islands is significantly better than the other \myapproach~with 5, 8, and 16 islands. No significant differences were found against \myapproach~with 21 island. Therefore, our results suggest the best approach is \myapproach~with 11 islands.

 Moreover, our results show that at some point increasing the number of islands do not generate benefits during the approximation of the Pareto optimal set. However, our results also suggest that larger island models can be useful for addressing bi-objective uniform instances.
 
\begin{figure}[t]
	 \centering
	 \begin{tabular}{c}
	 	\includegraphics[width=0.35\textwidth]{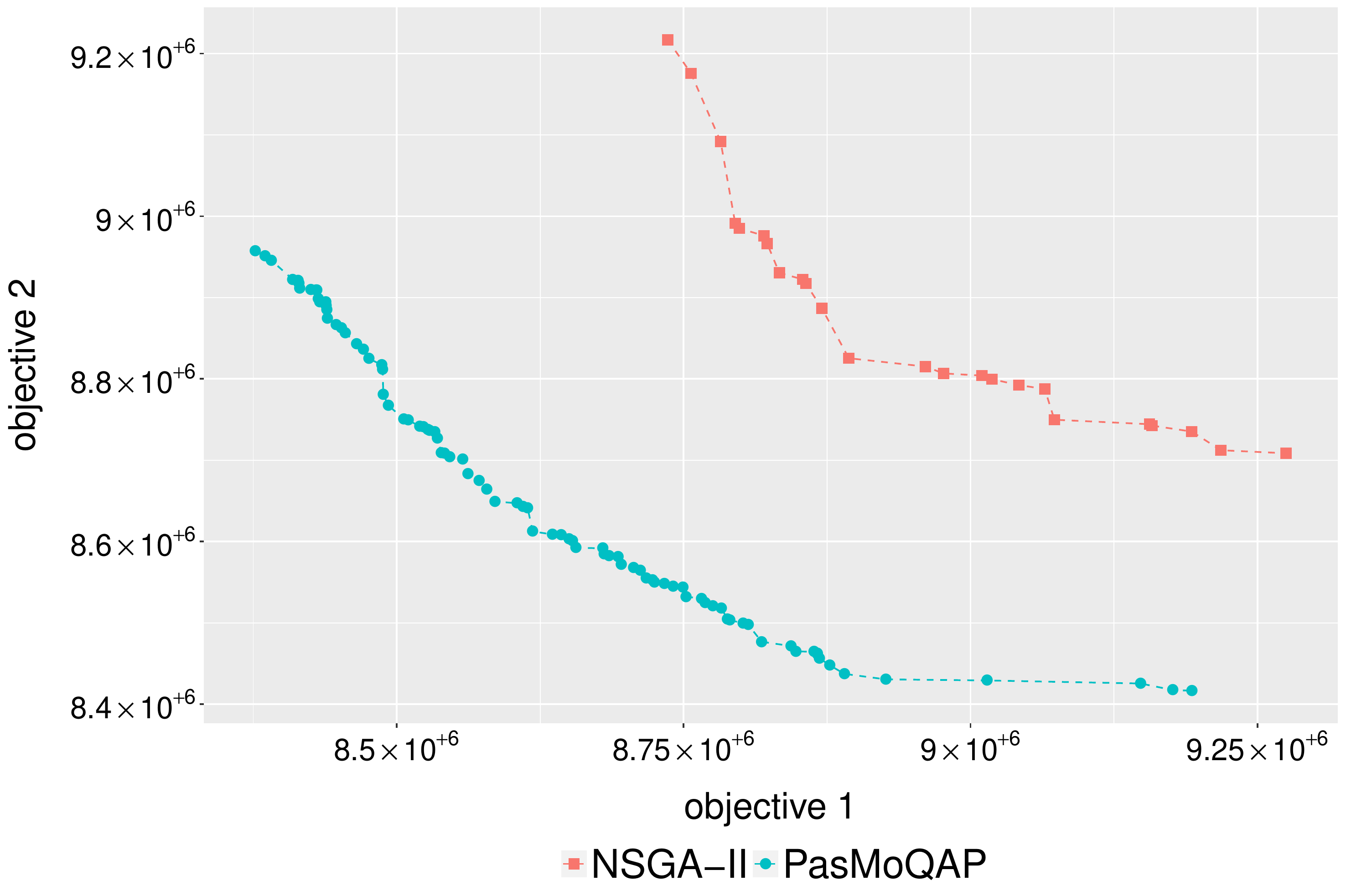} \\  \includegraphics[width=0.35\textwidth]{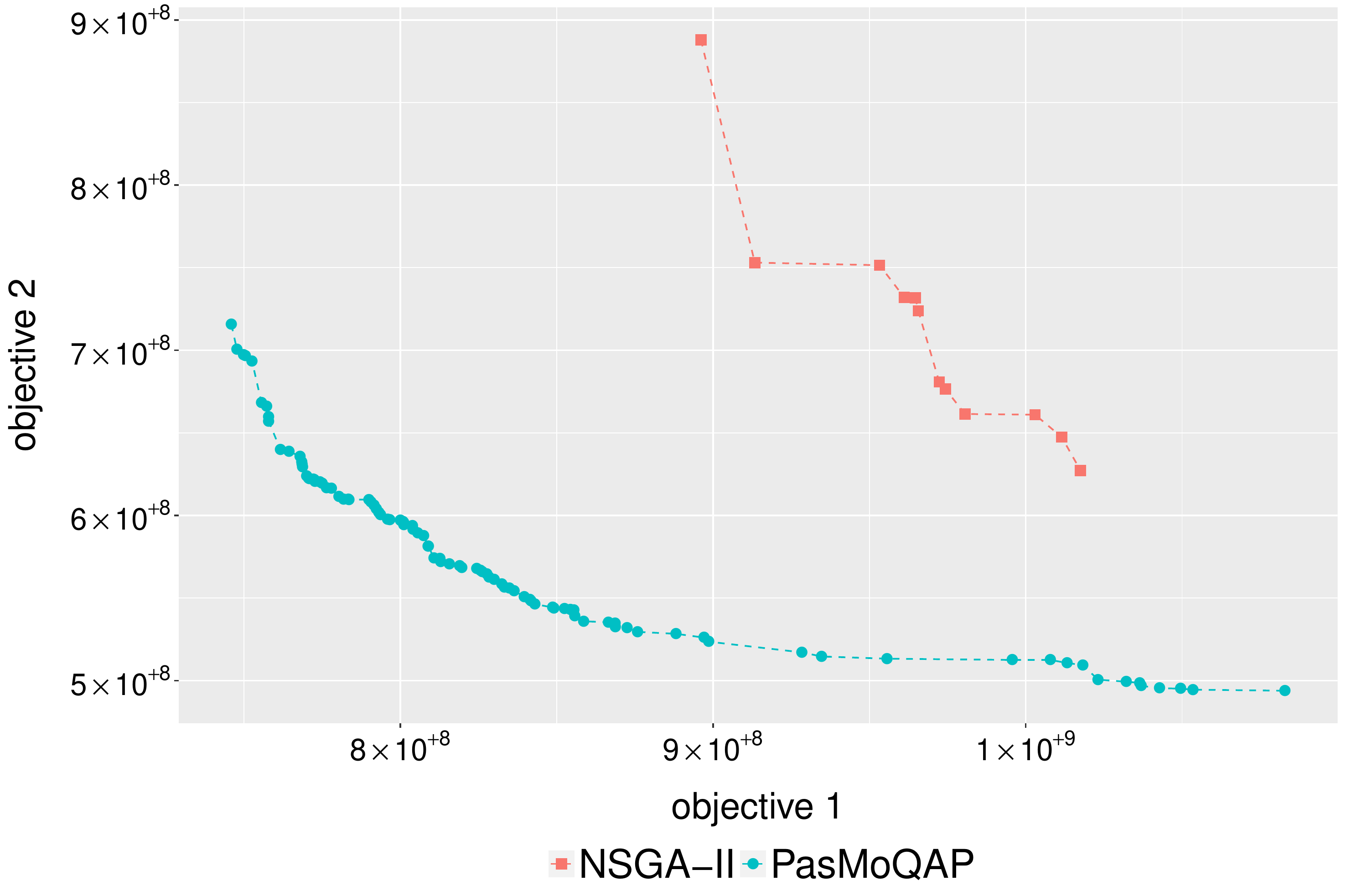} \\
	 \end{tabular}
	 \caption{The global Pareto fronts of the bi-objective uniform instance \instance{Gar60-2fl-1uni} (top) and the bi-objective real-like instance \instance{Gar60-2fl-1rl} (bottom). In both cases, \myapproach~computes better Pareto sets.}
	 \label{experiments:fig:pareto_fronts}
 \end{figure}

 We show in \autoref{experiments:fig:pareto_fronts} two Pareto fronts of bi-objective instances. We compute the Pareto fronts combining the 30 trials of the uniform instance \instance{Gar60-2fl-1uni} and the real-like instance \instance{Gar60-2fl-1rl}. We can see in both cases how \myapproach~computes better sets of non-dominated solutions.

 In summary, we observe that \myapproach~achieve a high performance compare to \acrshort{NSGATWO} in uniform and real-like instances using the hypervolume indicator. For this reason, \myapproach~is a promising alternative for solving \acrlong{mQAPs}.

\section{Conclusions and future work}
\label{sec:conclusions}

We propose the \myapproach~a parallel memetic algorithm for tackling the \acrlong{mQAP}. Our approach is based on the island model with small subpopulations. Each island evolves independent subpopulations, communicating asynchronously the promising solutions to the neighboring islands according to a user-defined topology. We preserve the diversity through external archives of limited size. We compare \myapproach~against a parallel version of the well-known multi-objective evolutionary algorithm \acrshort{NSGATWO}. The results show that \myapproach~significantly outperforms the baseline method. We observe that our proposal is a viable alternative for solving the \acrlong{mQAP} at a reasonable computational time. 

Further studies are required using more benchmark instances and more objectives (i.e. many objectives). We are interested in analyzing with more details the effects of the parameters involved in the island models (i.e. topologies, migration and integration techniques) that might give us more insights into the design of new parallel memetic algorithms for addressing the \acrshort{mQAP}.  Moreover, we believe that these parallel models might be improved if we explore the search space with different algorithms. These methods are called heterogeneous island models, and they might help to find a better balance between the algorithms' exploration and exploitation features. More empirical studies in this direction may help to identify advantages and disadvantages of each approach and how they can collaborate to approximate the optimal Pareto set.

\bibliographystyle{abbrv}
\bibliography{00_pasmoqap}  

\end{document}